%% file: main.tex
\documentclass[acmtog,authorversion,nonacm]{acmart}

\input{sections/_preamble.tex}
\input{sections/_template_args.tex}

\begin{document}

\title{Robust Dancer: Long-term 3D Dance Synthesis Using Unpaired Data}

\input{sections/0_authors.tex}

\input{sections/0_abstract.tex}
\input{sections/0_article_info.tex}


\maketitle

\input{sections/1_introduction.tex}
\input{sections/2_related_work.tex}
\input{sections/3_model.tex}
\input{sections/4_experiment_setup.tex}
\input{sections/5_experiment_results.tex}
\input{sections/6_conclusion.tex}

\bibliographystyle{ACM-Reference-Format}
\bibliography{main}

\input{sections/7_appendix.tex}

\end{document}

%% file: sections/_preamble.tex
\usepackage{bm}
\usepackage{bbm}
\usepackage{color}

\usepackage{graphicx}
\usepackage{tabularx}
\usepackage{subcaption}
\usepackage{multirow}
\usepackage{amsmath}
\usepackage{mathrsfs}
\newcommand{\vect}[1]{\bm{#1}}

\newcommand{\eqword}[1]{{\text{#1}}}

\usepackage{xspace}

\newcommand{\etal}{\textsl{et~al.}~}

\newcommand{\keep}[1]{}
\newcommand{\old}[1]{}

\usepackage{url}
\usepackage{hyperref}
\usepackage{eso-pic}
\newcommand{\fig}{Figure{}~}


%% file: sections/_template_args.tex
\AtBeginDocument{%
  \providecommand\BibTeX{{%
    \normalfont B\kern-0.5em{\scshape i\kern-0.25em b}\kern-0.8em\TeX}}}

\setcopyright{acmcopyright}
\acmJournal{TOG}
\acmYear{2022}
\acmVolume{0}
\acmNumber{0}
\acmArticle{0}
\acmMonth{0}
\acmDOI{x}

\setcopyright{none}

\acmSubmissionID{0}



%% file: sections/0_authors.tex
\author{Bin Feng}
\authornote{Both authors contributed equally to this research.}
\email{fengbin14@pku.edu.cn}
\affiliation{%
  \institution{National Key Laboratory for Multimedia Information Processing \& School of Computer Science, Peking University}
  \streetaddress{No.5 Yiheyuan Road, Haidian District}
  \city{Beijing}
  \state{Beijing}
  \country{China}
  \postcode{100871}
}

\author{Tenglong Ao}
\authornotemark[1]
\email{aubrey.tenglong.ao@gmail.com}
\affiliation{%
  \institution{Peking University}
  \streetaddress{No.5 Yiheyuan Road, Haidian District}
  \city{Beijing}
  \state{Beijing}
  \country{China}
  \postcode{100871}
}

\author{Zequn Liu}
\email{zequnliu@pku.edu.cn}
\affiliation{%
  \institution{National Key Laboratory for Multimedia Information Processing \& School of Computer Science, Peking University}
  \streetaddress{No.5 Yiheyuan Road, Haidian District}
  \city{Beijing}
  \state{Beijing}
  \country{China}
  \postcode{100871}
}

\author{Wei Ju}
\email{juwei@pku.edu.cn}
\affiliation{%
  \institution{National Key Laboratory for Multimedia Information Processing \& School of Computer Science, Peking University}
  \streetaddress{No.5 Yiheyuan Road, Haidian District}
  \city{Beijing}
  \state{Beijing}
  \country{China}
  \postcode{100871}
}

\author{Libin Liu}
\email{libin.liu@pku.edu.cn}
\affiliation{%
  \institution{Peking University}
  \streetaddress{No.5 Yiheyuan Road, Haidian District}
  \city{Beijing}
  \state{Beijing}
  \country{China}
  \postcode{100871}
}

\author{Ming Zhang}
\email{mzhang_cs@pku.edu.cn}
\affiliation{%
  \institution{National Key Laboratory for Multimedia Information Processing \& School of Computer Science, Peking University}
  \streetaddress{No.5 Yiheyuan Road, Haidian District}
  \city{Beijing}
  \state{Beijing}
  \country{China}
  \postcode{100871}
}

\renewcommand{\shortauthors}{Feng, Ao, Liu, Ju, Liu and Zhang}

%% file: sections/0_abstract.tex
\begin{abstract}
    How to automatically synthesize natural-looking dance movements based on a piece of music is an incrementally popular yet challenging task. Most existing data-driven approaches require hard-to-get paired training data and fail to generate long sequences of motion due to error accumulation of autoregressive structure. We present a novel 3D dance synthesis system that only needs unpaired data for training and could generate realistic long-term motions at the same time. For the unpaired data training, we explore the disentanglement of beat and style, and propose a Transformer-based model free of reliance upon paired data. For the synthesis of long-term motions, we devise a new long-history attention strategy. It first queries the long-history embedding through an attention computation and then explicitly fuses this embedding into the generation pipeline via multimodal adaptation gate (MAG). Objective and subjective evaluations show that our results are comparable to strong baseline methods, despite not requiring paired training data, and are robust when inferring long-term music. To our best knowledge, we are the first to achieve unpaired data training - an ability that enables to alleviate data limitations effectively. Our code is released on \href{https://github.com/BFeng14/RobustDancer}{https://github.com/BFeng14/RobustDancer}.
\end{abstract}

%% file: sections/0_article_info.tex
\ccsdesc[500]{Computing methodologies~Animation}
\ccsdesc[300]{Computing methodologies~Neural networks}

\keywords{Music-driven dance synthesis, multi-modality}

%% file: sections/1_introduction.tex
\section{Introduction}
\label{sec:introduction}
Dance is a full-body performing art consisting of rhythmic human movements. It is an important way for humans to inspire creativity and convey various emotions. There are strong demands for high-quality 3D dance animations in some industries, such as films, video games, and virtual avatars. Therefore, given a piece of music, automatically synthesizing natural-looking dance movements becomes a research hotspot recently.

Because of recent advances in deep learning, it could be possible to train a complex model using paired music-dance data, which achieves eye-catching results successfully \cite{alemi2017groovenet,li2021choreographer,chen2021choreomaster,lee2019dancing2music,siyao2022bailando,guillermo2021transflower}. These data-driven methods are supervised and rely on paired training data, where motions are strictly aligned with the music. However, collecting and aligning paired data is labor-intensive and expensive. Limited paired dataset hinders the capability of neural network. In contrast, unpaired data are easier to obtain, because music and motions are not required temporally frame-to-frame alignment. And so, it would be of great benefit if we can exploit the large amount of unpaired data to achieve model training.

Moreover, many existing systems fail to generate realistic long sequences of dance movements \cite{lee2018listentodance,ren2019poseperceptualloss,li2021choreographer,li2022danceformer}. Specifically, auto-regressive models like recurrent neural network (RNN) are widely applied to the dance synthesis \cite{lee2018listentodance,ren2019poseperceptualloss}. But the generated motions tend to be quickly stiff within a few seconds due to an incremental accumulation of prediction errors brought by the auto-regressive manner. Although recent developments of Transformer-based models \cite{li2021choreographer,li2022danceformer} show a powerful ability of cross-modal modeling, these systems generate dance based on short historical motions and ignore the long-term choreographic structure, resulting in inconsistent styles and rigid movements. In a practical context, we need an efficient mechanism to synthesize long-term dance motions robustly.

In this paper, we focus on the long-term 3D dance motion synthesis, accompanying with the music content on both rhythm and style levels. For the data-limited challenge above, we propose an efficient unpaired data training scheme based on the disentanglement of beats and styles of both music and motions. More specifically, we first divide the input interfaces of our system into beat and style two parts. Beat input could be identified from a given piece of music or a dance, while a Transformer-based style encoder is designed to extract style input by feeding style exemplars of music and motion. And we fuse the style input into the generator in a time-agnostic way via conditional layer normalization (CLN) \cite{kexuefm-7124}. During training, we obtain beats from the ground-truth dance instead of music, while the style exemplars of music and motion are randomly sampled pieces, which are consistent with the style label of the ground-truth dance. The whole scheme does not require any paired data, and it only requires the easy-to-get style labels. As for the inference, we identify beats from the given music to replace the motion beats used during training, and the given music clip and a fixed style-consistent dance are fed into the style encoder as the style exemplars. Both quantitative and qualitative evaluations show that our system performs comparably to strong baseline methods that use paired data.

For generating natural-looking long-term dance movements, we develop a novel long-history attention strategy to capture and maintain a choreographic structure, which effectively alleviates error accumulation of auto-regressive manner. The whole strategy includes two phases. We first use the latest motions to query a long-range historical dance via a newly designed attention computation and get a long-history embedding. Then, the embedding is fused into the generation pipeline via multimodal adaptation gate (MAG) \cite{rahman2020integrating}, explicitly give a historical hint to the synthesis of motions. Experiments demonstrate that our system robustly synthesize realistic long-term dance movements.

In summary, our contributions are three-folds:
\begin{itemize}
    \item We develop a novel 3D dance synthesis system that can robustly generate impressive dances accompanying with a piece of long music. To our best knowledge, it is the first system that successfully uses only unpaired data to achieve music-driven dance generation.
    \item We propose an efficient unpaired data training scheme to alleviate the problem of lacking data.
    \item We devise an innovative long-history attention strategy, which explicitly maintains the temporal coherence between music and dance in the long-term time.
\end{itemize}

%% file: sections/2_related_work.tex
\section{Related Work}
\label{sec:related_work}

\subsection{Human Motion Prediction}
The goal of the human motion prediction task is to predict future motions based on historical motions, which requires the model to learn a robust prior of human movements, including the skeleton structure and the underlying kinematic information. Conventional methods \cite{arikan2002interactive,kim2003rhythmic} usually construct a motion graph \cite{kovar2008motiongraph} and find an optimal path based on the transition probabilities between nodes. Motion graph has good interpretability, but the motion transitions are usually not smooth and the diversity of generated motions is limited by the graph scale. With the development of deep learning, deep-based methods have dominated this area. Compared with motion graphs, neural networks are able to generate smooth and diverse motions while the model can effectively compress large-scale dataset into the network parameters. Most existing deep-based models opt an auto-regressive manner to model the prediction of human motions, like recurrent neural networks (RNNs) \cite{fragkiadaki2015recurrent,mahmood2019amass,ghosh2017learning,martinez2017human,li2017auto,ao2022rhythmicgesticulator}, convolutional neural networks (CNNs) \cite{li2018convolutional,liu2020trajectorycnn}, and recent Transformer-based networks \cite{aksan2021spatio,mao2022weakly,ao2023gesturediffuclip}.

\subsection{Music-driven Dance Synthesis}
Music-driven dance synthesis system aims to model the complex temporal harmony between music and dance. Existing methods are divided into two lines, which are the motion graph-based method and the end-to-end method.

\subsubsection{Motion Graph-based Method}
The core of the motion graph-based method is to model the matches between two motion sequences and between music and motion. Early methods use some statistical models like the hidden markov model (HMM) \cite{kim2006making,lee2013music} to compute the transition probability between motion sequences and the emission probability between music and motion. Recently some works \cite{chen2021choreomaster,ye2020choreonet} explore the utilization of deep neural networks to model the matches above and achieve convincing results on the carefully collected dataset. Although motion graph-based methods can generate realistic motions, they must trade off the model capacity and the complexity of inference because of fussy training procedures.

\subsubsection{End-to-end Method}
Due to the strict requirement of the dataset and complicated training procedures, the popularity of motion graph-based methods is fading and end-to-end methods dominate the dance synthesis. In the context of the end-to-end model, dance generation is defined as a sequence-to-sequence translation task from music to motions without carefully labeled datasets. One line of the end-to-end model is the deterministic model. Different model structures, such as restricted boltzmann machines \cite{alemi2017groovenet}, LSTM \cite{huang2020dance,aristidou2021rhythm}, and Transformer \cite{li2021choreographer,li2022danceformer,siyao2022bailando}, are tried to improve the synthesis performance. Another line belongs to the probabilistic model. Because of the inherent one-to-many mapping from music to dance, some works utilize probabilistic models like generative adversarial networks (GANs) \cite{lee2019dancing2music,sun2020deepdance,ferreira2021learning} and normalizing flow \cite{guillermo2021transflower} to avoid falling into a mean pose. Most end-to-end approaches follow the auto-regressive manner, which suffers from severe error accumulation problem in long motion sequence generation. Both Huang \etal \cite{huang2020dance} and Aristidou \etal \cite{aristidou2021rhythm} draw on the idea of acRNN \cite{li2017auto} to alleviate this problem. Because we use Transformer instead of RNN, we design a new attention mechanism to avoid generated motions to freeze during long-term dance synthesis.

Most of the existing methods, whether motion graph-based or end-to-end, require paired data for training, which suffers from the data-limited problem. In this paper, we propose an unpaired data training scheme to loose the requirement on the dataset.

%% file: sections/3_model.tex
\section{Problem Definition}
\label{sec:definition}
The dance generation problem is defined as an autoregressive manner. Distinguishing from previous works \cite{huang2020dance,li2021choreographer} that directly took the music context as input, we decouple the music context into music style and music beat. More precisely, our model takes three inputs: 1) \emph{historical motion}, which is a sequence of poses $\vect{X}_{1:T}=\{{\vect{x}_{1}}, \cdots, {\vect{x}_{T}}\} \in \mathbb{R}^{T\times(J\times9+3)}$, where $J$ is joint amount with the representation of rotation matrix and the root trajectory is also included; 2) \emph{beat context}, which means a longer sequence of binary value $\vect{B}_{1:T+n}=\{b_{1}, \cdots, b_{T+n}\}$, where $b_{i} \in \{0,1\}$, and $n > 0$; 3) \emph{style exemplars}, which guide the style of generated dance and are represented as fixed $w$-length sequences of music feature $\vect{M}$ and motion $\vect{X}'$, where the music feature consists of 20-D MFCCs and 12-D chorma features. Then, the task is to predict the next pose ${\vect{x}_{T+1}}$ given historical motion, beat context, and style exemplars. To alleviate the error accumulation brought by autoregressive structure, similar to \cite{li2021learn}, we let the model predict future $n$ frames instead of single next frame at each time during training, and we set $n=7$ in our experiments.

\section{Methodology}
\label{sec:model}

\subsection{Model Overview}
\begin{figure}[t]
    \centering
    \includegraphics[width=0.9\linewidth]{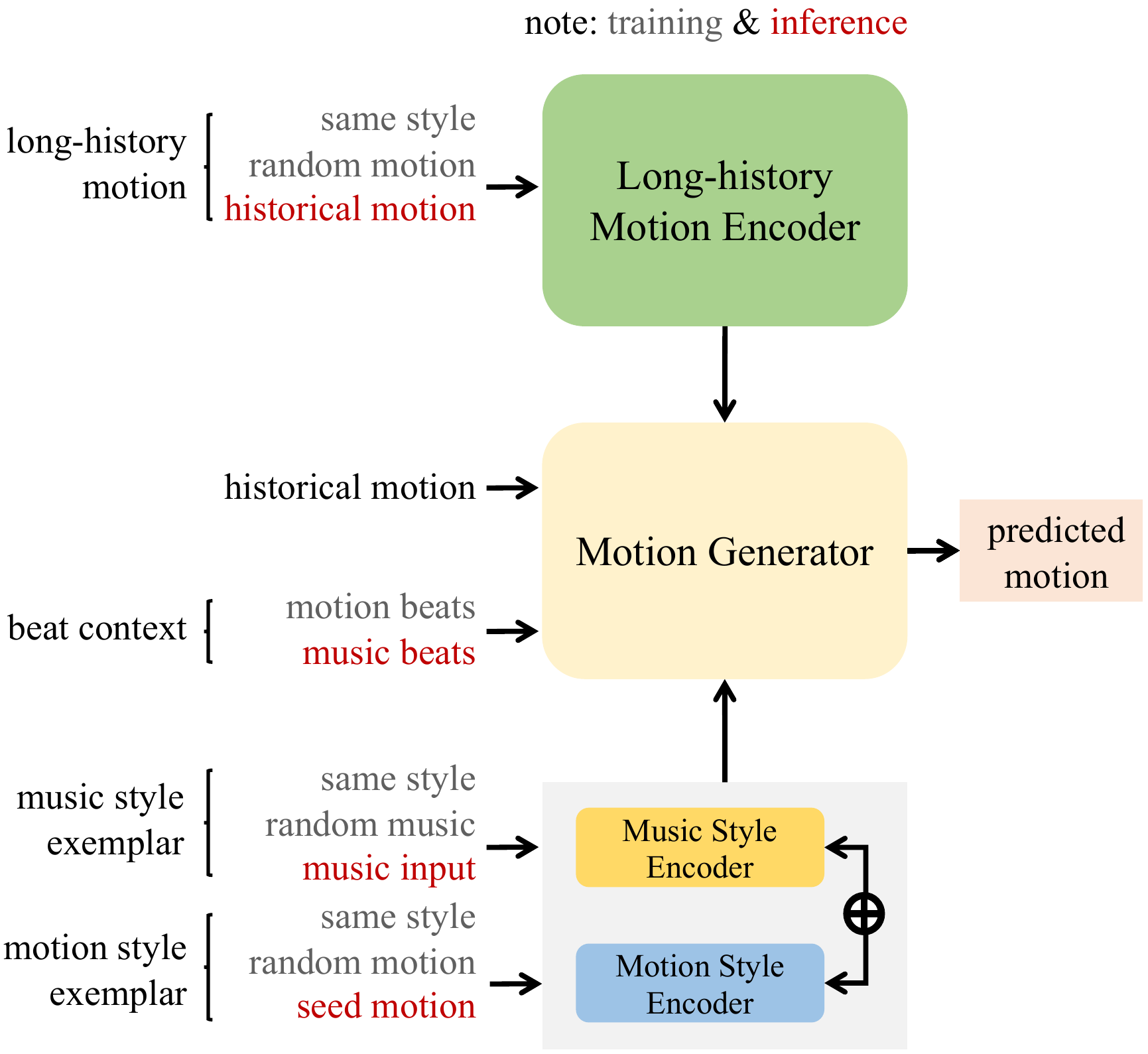} 
    \caption{\textbf{Model Overview.} Our model is composed of three core components: (a) the \emph{style encoder} integrates the style information from the music and motion style exemplars, enabling that the generated motions have consistent style with the seed motion and the input music, then (b) the \emph{long-history motion encoder} encodes long historical motion into a embedding, which carries choreographic structure information to guide the synthesis of long-term motion, (c) the \emph{motion generator} fuses the style information and long-history embedding into the pipeline, and performs the beat-guided motion generation autoregressively. The entire training process requires only unpaired datasets.}
    \label{fig:overview}
\end{figure}
Paired data refers to music-dance sequential pairs that are strictly aligned frame by frame. In contrast, unpaired data means separate, non-temporally aligned music and dance sequences. Existing methods \cite{lee2018listentodance,li2021choreographer,guillermo2021transflower} can only use paired datasets for training, which are expensive to obtain. And these end-to-end models trained on paired data tend to over-focus on low-level details of frame-to-frame alignment between different modalities, while ignoring the more important high-level features of choreographic rhythm and style. Then they are prone to overfitting. For a well-trained human choreographer, he can choreograph a natural dance even if only the rhythmic structure (e.g., beat) and dance style are specified. A good generator needs to have this top-down generation capability. To learn a robust generator, we imitate this choreography process, while taking advantage of the easy-to-get unpaired data.

As illustrated in \fig\ref{fig:overview}, like the inputs for the human choreographer, we decouple the input context into the beat context and the style exemplars. The beat context controls the rhythmic structure, and the style exemplars specify the style of generated movements by feeding into the style encoder. When training, to avoid utilizing the paired data, we identify the motion beats as the beat context, while we randomly sample the music and motion segments from the dataset according to the style label of the target dance motion as the style exemplars. This random sampling introduces reasonable noise to the training, which enhances the model robustness. Then during inference, we directly use the input music to identify the beat context and music style exemplar respectively. A fixed seed motion as the motion style exemplar gives a style hint to the generator. The whole process does not require any paired data.

We perform the motion generation in an autoregressive manner, which tends to be style-inconsistent when synthesizing a long-term dance due to the accumulation of prediction errors. To generate realistic long-term dance, except for employing techniques like predicting multiple frames at a time, explicit modeling of choreographic structure has proven effective in previous works \cite{chen2021choreomaster,aristidou2021rhythm}. The motion repeat constraint is one of the most common rules in choreography theory. Choreographers often utilize repeat motion to echo the repeated phrases (e.g., verse and chorus) in music \cite{chen2021choreomaster}. To include this constraint, we design a new long-history motion encoder, which can represent the ongoing motion as a query to scan a similar motion window from the long historical motion. Then, we fuse the queried structural information into the generation pipeline, which guides the generator to synthesize realistic long-term motion in a choreographic-aware way. A similar noisy training strategy (see \fig\ref{fig:overview}) is also introduced to make model robust.

\subsection{Style Encoder}
We use Transformer \cite{vaswani2017attention}, which has been widely used to model multi-modal data in many successful temporal systems \cite{huang2020dance,li2021choreographer,siyao2022bailando,fan2022faceformer}, to encode the music and motion style exemplars respectively.
The music style encoder $\mathcal{E}_M$ encodes the music style exemplar $\vect{M}$ into the music style embedding matrix $\vect{H}_{\text {music}}$, while the motion style encoder $\mathcal{E}_X$ encodes the motion style exemplar $\vect{X}'$ into the motion style embedding matrix $\vect{H}_{\text {motion}}$. Note that both music and motion style encoders have the same network structure, so the shapes of output matrices are both $w \times d_s$, and we set the length of examplar $w$ to 2 seconds in this paper. Finally, we get the integrated style embedding matrix $\vect{H}_{\text {style}}$ by adding $\vect{H}_{\text {music}}$ to $\vect{H}_{\text {motion}}$.

\subsection{Long-history Motion Encoder}
\begin{figure}[t]
    \centering
    \includegraphics[width=\linewidth]{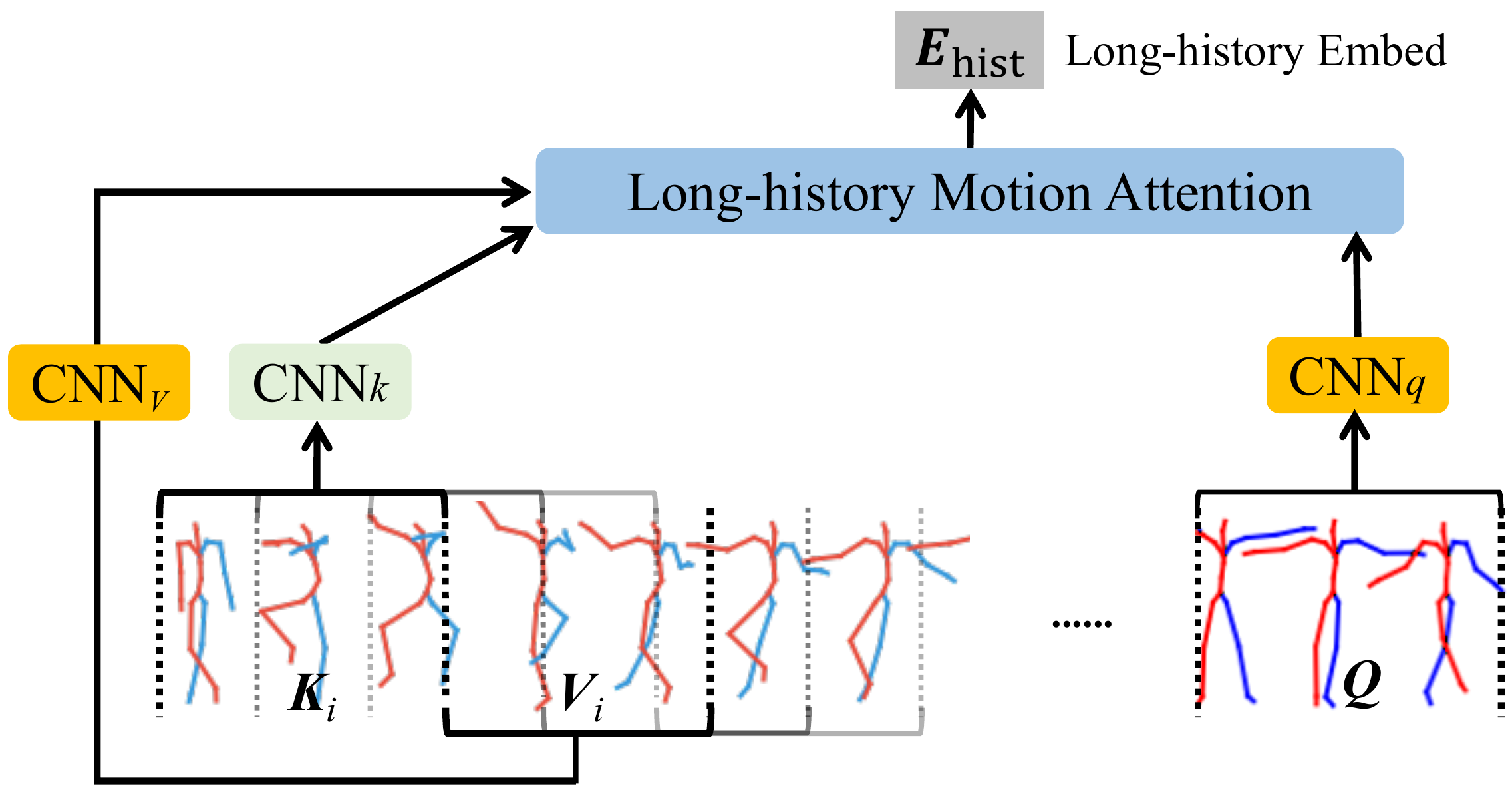} 
    \caption{\textbf{Long-history Motion Encoder.} We use the ongoing motion (query) to find a similar historical motion window (key), then the subsequent motion (value) of the motion window is helpful for the prediction of future motion according to the motion repeat constraint in choreographic theory.}
    \label{fig:long_hist}
\end{figure}
Attention mechanism is naturally suited for mining useful information from motion sequences \cite{mao2020history,jang2022motionpuzzle}. We present an attention mechanism to fit the motion repeat constraint in choreographic theory. We assume that if the ongoing motion is very similar to a historical motion window, then the future motion is likely to be similar to the subsequent motion of the motion window. As illustrated in \fig\ref{fig:long_hist}, we use the $m$-frames motion window ${\vect{X}}_{i: i+m-1}$ as key, the $n$-frames subsequent motion of the history motion ${\vect{X}}_{i+m: i+m+n-1}$ as value, and the ongoing $m$-frames motion ${\vect{X}}_{T-m+1: T}$ as the query.
\begin{align}
    {\vect{q}} &= \eqword{AvgPool}(\eqword{CNN}_{q}\left({\vect{X}}_{T-m+1: T}\right)) \in \mathbb{R}^{d}, \\
    {\vect{k}_i} &= \eqword{AvgPool}(\eqword{CNN}_{k}\left({\vect{X}}_{i: i+m-1}\right)) \in \mathbb{R}^{d}, \\
    {\vect{V}_i} &= \eqword{CNN}_{V}({\vect{X}}_{i+m: i+m+n-1}) \in \mathbb{R}^{n\times d},
\end{align} 
\noindent
where $i \in\{1,\cdots, T-m-n+1\}$, $\eqword{CNN}(\cdot)$ means 1-D convolutional neural network, $\eqword{AvgPool}(\cdot)$ means average pooling, and we set window size $m=10$ in this work. After obtaining the query, key, and value, we perform the standard dot-product attention calculation as
\begin{equation}
    a_{i}=\frac{\eqword{exp}({\vect{q} \vect{k}}_{i}^{T})}{\sum_{i=1}^{T-m-n+1} \eqword{exp}({\vect{q}} {\vect{k}}_{i}^{T})}.
\end{equation}
Then we get the long-history embedding matrix
\begin{equation}
    {\vect{E}_{\eqword{hist}}}=\sum_{i=1}^{T-m-n+1} a_{i}\vect{V}_i \in \mathbb{R}^{n \times d}.
\end{equation}
Moreover, we use randomly sampled motion segments from the dataset, which keep the same style label as the ground-truth dance, to replace the historical motion during training (see \fig\ref{fig:overview}). The noisy training mechanism helps to enhance the model generalization.

\subsection{Motion Generator}
\begin{figure*}[t]
    \centering
    \includegraphics[width=0.9\textwidth]{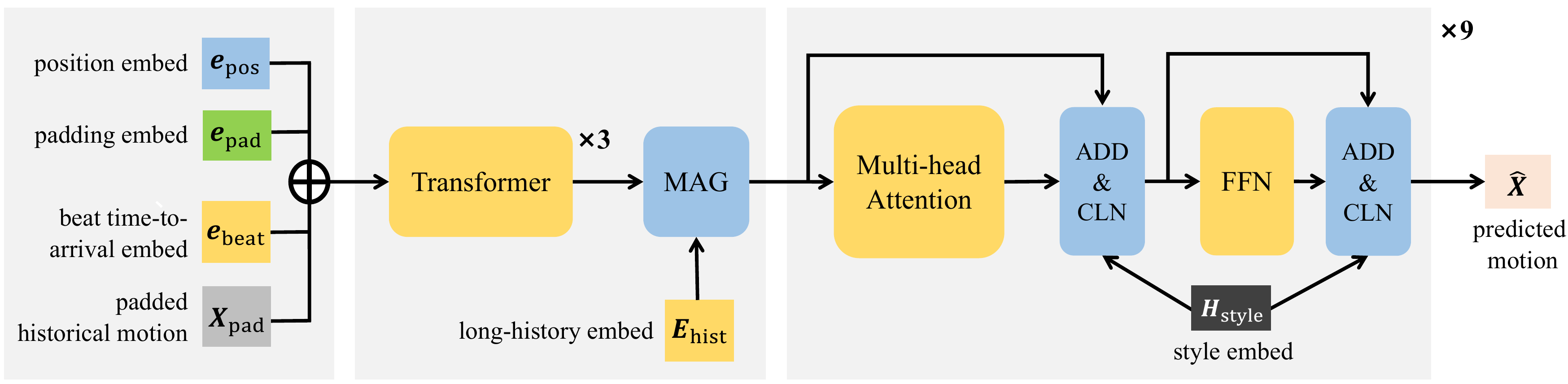} 
    \caption{\textbf{Motion Generator.} The Transformer-based generator synthesizes a realistic dance sequence, conditioned on the historical motion, the beat context, the long-history embedding, as well as the style embedding.}
    \label{fig:generator}
\end{figure*}

Given the latest $w$ frames of historical motion $\vect{X}_{T-w+1:T}$, beat context $\vect{B}_{T-w+1:T+n}$, long-history embedding ${\vect{E}_{\text {hist}}}$, and style embedding ${\vect{H}_{\text {style}}}$, we use a Transformer-based model $\mathcal{G}$ to synthesize future $n$ frames of dance movements
\begin{align}
    \hat{\vect{X}}_{T+1:T+n} = \mathcal{G}(&\vect{X}_{T-w+1:T}, \vect{B}_{T-w+1:T+n}, \vect{E}_{\text {hist}}, \vect{H}_{\text {style}}).
\end{align}
\noindent
We perform some pre-processing on the inputs before feeding them into the Transformer. As shown in \fig\ref{fig:generator}, the input historical motion is padded by repeating the last frame $n$ times $\vect{X}_{\text{pad}} \in \mathbb{R}^{(w+n) \times (J\times9+3)}$. On the one hand, the padding operation adapts the historical motion length to the Transformer output. On the other hand, due to the smoothness of movements, the next few frames of motion are usually close to the $\vect{x}_T$, then our padding operation can reduce the learning difficulty by letting the model learn the residuals. Moreover, we add the padding embedding $\vect{e}_{\text{pad}}$, which is a binary vector to indicate whether the current position is padded or not, together with the standard position embedding $\vect{e}_{\text{pos}}$ into the inputs.

\subsubsection{Style and Beat Fusion.}
We consider that music and dance are temporally aligned in beat and globally matched in style. Then our high-level idea is to fuse the style embedding in a time-agnostic way, while fusing the beat context in a time-aware way.

To modulate the generated dance with the style globally, we apply the conditional layer normalization (CLN) \cite{kexuefm-7124} to replace the standard layer normalization in Transformer. The key idea of CLN is to predict the affine coefficients $\gamma$ and $\beta$ from the style embedding, which manipulate feature maps by scaling and shifting different channels. \citet{kexuefm-7124} found that it is not easy to directly predict these two coefficients, as it may make the training process unstable. To solve this, we choose to predict the changes $\Delta \gamma$ and $\Delta \beta$ using the one-hidden-layer MLP, 
\begin{align}
    \Delta \beta = \operatorname{MLP_{\beta}}\left(\vect{H}_{\text {style}}\right),
    \Delta \gamma = \operatorname{MLP_{\gamma}}\left(\vect{H}_{\text {style}}\right),
\end{align}
\noindent
which makes prediction easier. Then, the style fusion can be formulated as
\begin{equation}
y=\frac{x-\mathbb{E}[x]}{\sqrt{\operatorname{Var}[x]+\epsilon}} \cdot (\gamma+\Delta \gamma) + (\beta +\Delta \beta).
\label{con:cln}
\end{equation}

To let the generator perceive the rhythmic structure during the generation, we include the time-to-arrival embedding $\vect{e}_{\text{beat}}$ to encode the temporal beat context. We translate each element of the binary beat sequence into the number of frames remaining until the next beat point, which helps the generator to learn a natural transition between beats. 

\subsubsection{Long-history Embedding Fusion.}
Long-history embedding carries memory about the choreographic structure from historical motion, and we need a gate mechanism similar to LSTM \cite{hochreiter1997lstm} to control the effects brought by the long-history memory. The Multimodal-Adaptation-Gate (MAG) method \cite{rahman2020integrating} was proposed for multimodal fusion via an adaptive gate. We apply the MAG to fuse the long-history embedding $\vect{E}_{\text{hist}}$ after the third layer of Transformer. More specifically, we first compute the gate factor
\begin{equation}
    \vect{g} = \mathcal{R}\left(\vect{W}_g\left[\vect{z}_{i} ; \vect{e}_{i}\right]+b_g\right),
\end{equation}
\noindent
which is used to inhibit unreliable memory. $\vect{z}_{i}$ is the output feature of the Transformer in the $i$-th temporal position, $\vect{e}_i$ denotes the $i$-th row of $\vect{E}_{\text{hist}}$, and $\mathcal{R}(\cdot)$ means the activation function. Then, we get the valid memory gated by $\vect{g}$
\begin{equation}
    \vect{h}_{i} = \vect{g} \odot \left(\vect{W}_h \vect{e}_{i}\right)+b_h,
\end{equation}
\noindent
where $\vect{W}_g, \vect{W}_h$ are weight matrices and $b_g, b_h$ are scalar biases. Finally, the fused feature $\bar{\vect{z}}_{i}$ is formulated as
\begin{align}
    \bar{\vect{z}}_{i}=\vect{z}_{i}+\alpha \vect{h}_{i},
    \alpha=\min \left(\frac{\left\|\vect{z}_{i}\right\|_{2}}{\left\|\vect{h}_{i}\right\|_{2}} \beta, 1\right),
\end{align}
\noindent
where $\beta$ is a hyper-parameter, and the scaling factor $\alpha$ controls the effect of $\vect{h}_i$ within a desirable range.

\subsection{Unpaired Learning Scheme}
\label{subsec:unpaired}
As shown in \fig\ref{fig:overview}, during training, we use the random music and motion, which have the same style label as the target dance, as the style exemplars, while identifying the ground-truth motion beats as the beat context. We perform the algorithm proposed by \cite{ho2013extraction} for the motion beat identification. The core idea of \cite{ho2013extraction} is to detect the local minima of joints deceleration as the motion beat. Moreover, to improve the clustering of the different styles in the latent space, we apply a triplet loss on the music style embedding space via
\begin{align}
    \mathcal{L}_\eqword{trip} = & \max \big\{\left\|(\mathcal{E}_M(\vect{M}), \mathcal{E}_M(\vect{M}_\eqword{pos})\right\|_{2}-\left\|\mathcal{E}_M(\vect{M}), \mathcal{E}_M(\vect{M}_\eqword{neg})\right\|_{2} + \delta, 0\big\},
\end{align}
\noindent
where $\vect{M}_\eqword{pos}$ denotes the positive sample that is a randomly sampled music clip of the same style as $\vect{M}$, while $\vect{M}_\eqword{neg}$ denotes the negative sample that is a randomly sampled music clip of the different style as $\vect{M}$. $\delta$ is the margin. The triplet loss encourages style embedding of the same cluster to be closer to each other. Finally, when inference, we identify the onsets of input music as the beat context to replace the motion beats used during training, and directly use the input music and fixed seed motion as the style exemplars.

We use a pose reconstruction loss for the predicted future $n$-frames motion, which is calculated by the L2 loss between the predicted motion and target motion:
\begin{equation}
    \mathcal{L}_{\eqword{rec}} = \sum\left\|\hat{\vect{x}}_{t}-\vect{x}_{t}\right\|_{2},
\end{equation}
\noindent
where $\hat{\vect{x}}_{t}$ is the predicted motion and $\vect{x}_{t}$ is the target motion. Again, we use a strategy of future-$n$ supervision to drive the model to pay more attention to the more temporal context during training by predicting future $n$ frames, and we keep only the first frame of the output as the predicted pose during the inference process.

To solve the human foot sliding problem, our model predicts the foot contact labels and use L2 loss between predicted and real foot contact labels to compute the foot contact loss. 
\begin{equation}
    \mathcal{L}_{\eqword{foot}} = \sum\left\|\hat{c}_{t}-c_{t}\right\|_{2},
\end{equation}
where $c_{t}, \hat{c}_{t}$ denote the true and predicted foot contact labels respectively. We extract the foot contact label by simply setting a foot velocity threshold.

Collectively, the total loss function is: 
\begin{equation}
    \mathcal{L} = \lambda_{\eqword{rec}}\mathcal{L}_{\eqword{rec}} + \lambda_{\eqword{foot}}\mathcal{L}_{\eqword{foot}} + \lambda_{\eqword{trip}}\mathcal{L}_{\eqword{trip}}
\end{equation}

%% file: sections/4_experiment_setup.tex
\section{Experiment Setup}
\label{sec:experiment_setup}
\subsection{Dataset}
We conducted experiments on a large-scale 3D human dance motion dataset AIST++ \cite{li2021learn}. It contains $1408$ dance sequences with a total duration of $18694$ seconds, and 10 dance genres (e.g., pop and street jazz). For motion data, we converted the SMPL representation to $219$-D motion features, including the flattened rotation matrix of $24$ body key points and the position of the root. We sampled motions at $20$ frames per second. Moreover, we adopted mirror motion schemes for data augmentation.

\subsection{Implementation Details}
The experiments were conducted on a TitanX GPU. In the training process, we set the batch size to $128$ and used the Adam optimizer \cite{kingma2014adam}. The model was trained for $100$k iterations, and the learning rate was set to $1e$-$4$ at the beginning of training, and decayed to $1e$-$5$ and $1e$-$6$ after $35$k and $60$k iterations, respectively. All Transformers have $10$ attention heads with $640$ hidden size, the dimension of hidden layers of FNN is $1920$. For $\mathcal{E}_M$ and $\mathcal{E}_X$, the number of attention layers is $3$. For $\mathcal{G}$, the number of attention layers is $12$, and a $2$-layer MLP with $1920$ hidden layer dimension is used in the conditional layer normalization. The dropout rate is set to $0.1$.

\subsection{Evaluation Metrics}
\begin{figure*}[t]
    \centering
    \includegraphics[width=0.9\textwidth]{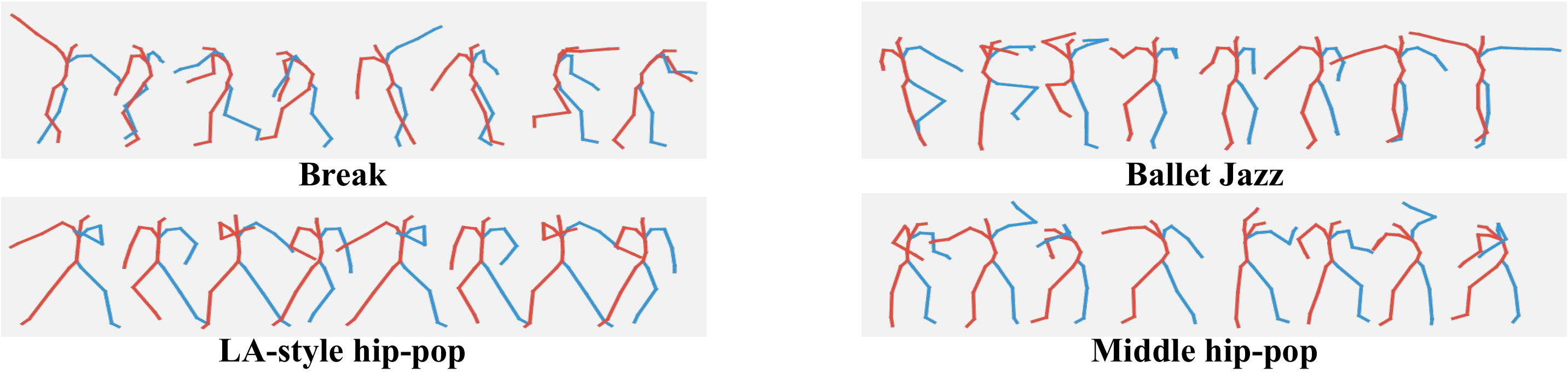} 
    \caption{\textbf{Dance Examples.} The dance motion synthesized by our model on different styles of music. }
    \label{fig:demo}
\end{figure*}
\begin{table*}[t]
    \centering
    \caption{\textbf{Quantitative and Qualitative Results on AIST++ Test Set.} Comparison of our method to DanceRevolution \cite{huang2020dance} and FACT \cite{li2021learn}. AIST++(random) means random-paired motion-music data.}
    \label{tab:analysis_real}
    \begin{tabular}{ccccccccc} 
        \hline
         & \multicolumn{2}{c}{Quality} & \multicolumn{2}{c}{Diversity} & \multicolumn{2}{c}{Motion-Music Corr} &
         \multicolumn{2}{c}{User Study} \\
         &  $\text{FID}_{k} \downarrow$ & $\text{FID}_{g} \downarrow$ & $\text{Dist}_{k} \uparrow$ & $\text{Dist}_{g} \uparrow$ &  $\text{BeatAlign}\uparrow$ & $\text{Acc}_S\uparrow$
         & Human Score$\uparrow$ & Ours WinRate \\
        \hline
        AIST++ & $-$ & $-$ & 7.90 & 8.21 & 0.278 & 0.67 & $4.23 \pm 0.18$ & 4.3\% \\
        AIST++(random) & $-$ & $-$ & 7.90 & 8.21 & 0.194 & - & $3.42 \pm 0.46$ & 35.2\% \\
        \hline
        DanceRevolution & 41.51 & 22.21 & 4.43 & 4.16 & 0.204 & 0.31 & $2.59\pm0.17$ & 78.5\% \\
        FACT & 38.13 & 20.31 & 5.78 & 6.18 & 0.222 & 0.35 & $2.36\pm0.35$ & 81.0\% \\
        Ours & \textbf{25.10} & \textbf{15.63} & \textbf{6.81} & \textbf{7.04} & \textbf{0.223} & \textbf{0.50} & \textbf{3.03} $\pm$ \textbf{0.29} & $-$\\
        \hline
    \end{tabular}
\end{table*}

\subsubsection{Motion Quality}
Following \cite{li2021learn}, we calculated the kinetic and geometric Fr{\'e}chet Inception Distances (FIDs) between generated motion and real motion to evaluate the motion quality. The corresponding FIDs are denoted as ${\eqword{FID}_k}$ and ${\eqword{FID}_g}$, respectively. 
    
\subsubsection{Motion Diversity}
To measure the diversity of generated motion, we adopted the metric proposed by \cite{li2021learn} to calculate the average Euclidean distance of the kinetic and geometric features, denoted as $\eqword{Dist}_k$ and $\eqword{Dist}_g$, respectively. 
    
\subsubsection{Beat Alignment}
We followed \cite{li2021learn} to use beat alignment score to evaluate how the generated motion align with the music beats. It is the average distance between each kinematic beat and its nearest music beat:
\begin{equation}
    \eqword{BeatAlign}=\frac{1}{m} \sum_{i=1}^{m} \exp \left(-\frac{\min _{\forall t_{j}^{y} \in B^{y}}\left\|t_{i}^{x}-t_{j}^{y}\right\|^{2}}{2 \sigma^{2}}\right),
\end{equation}
where $\left\{t_{i}^{x}\right\}$ and $\left\{t_{j}^{y}\right\}$ are kinematic and music beats respectively. The normalization parameter $\sigma$ is set to $1$.

\subsubsection{Style Matching}
We trained a SVM classifier on AIST++ to measure whether the style of generated dance matches the input music. The classifier takes extracted kinetic and geometric features as inputs, and we calculate the classification accuracy $\eqword{Acc}_S$ as the metric.

%% file: sections/5_experiment_results.tex
\section{Results}
\label{sec:experiment_results}
\fig\ref{fig:demo} shows the dance synthesis results for the music pieces from the AIST++ test set. Different styles of realistic dance movements were generated by our method.

We compared our model with two strong baselines. \emph{DanceRevolution} \cite{huang2020dance} is a seq2seq-based dance generation model with a curriculum learning strategy to alleviate error accumulation during long-term motion generation. \emph{FACT} \cite{li2021learn} is a full-attention cross-modal Transformer. Both baselines are reproduced using official released code with the optimal settings.

\subsection{Quantitative Evaluation}
Table \ref{tab:analysis_real} summarizes the objective results. Despite not using paired data, our method still achieves the lowest FID valued. The diversity of our generated motion is comparable to the strong baseline methods. The style classification accuracy of our model reaches $50\%$, which significantly exceeds strong baseline methods by $15\%$.

\subsection{Long-term Dance Synthesis}
\begin{figure}[t]
    \centering
    \includegraphics[width=0.9\linewidth]{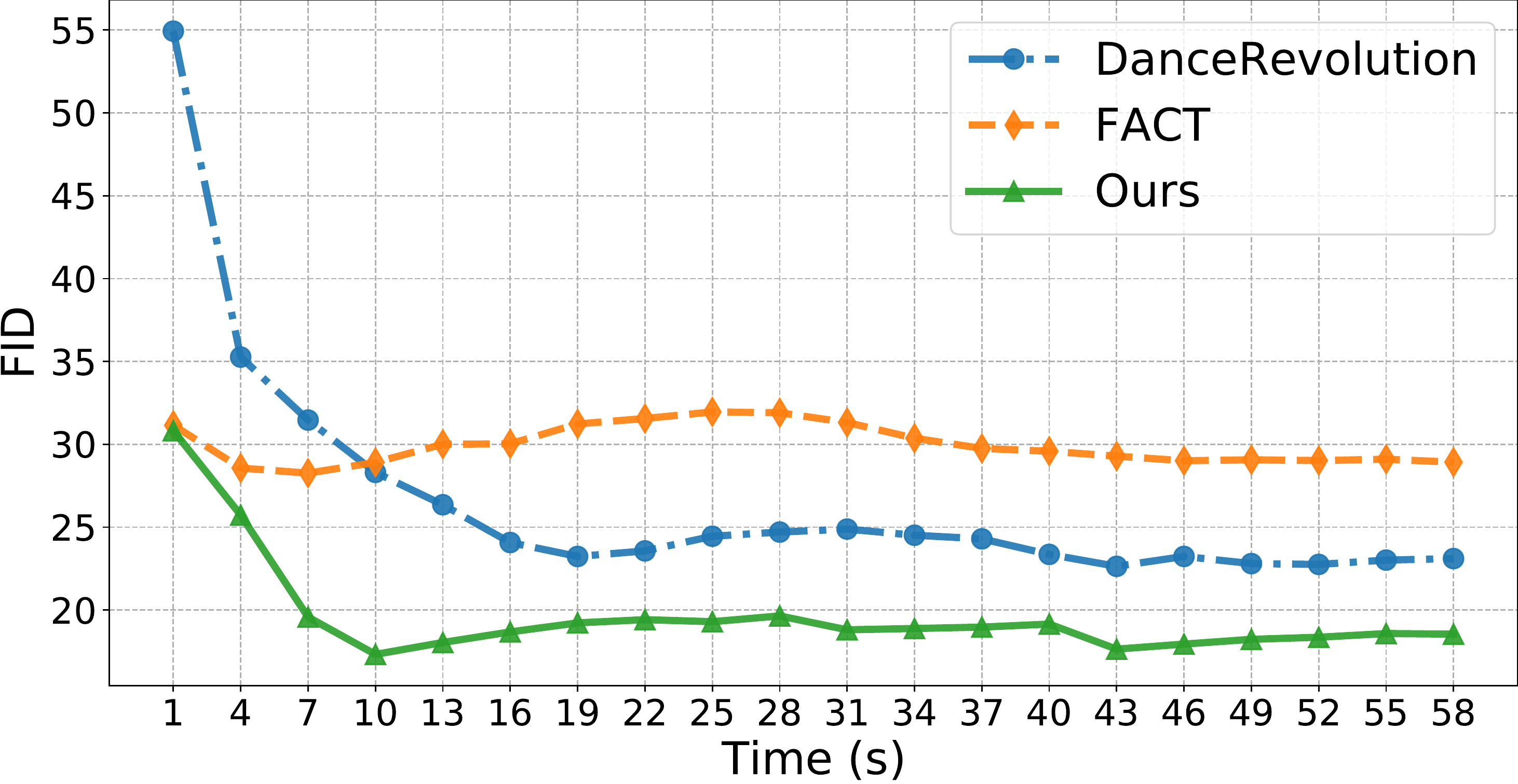} 
    \caption{\textbf{Experiment of Long-term Dance Synthesis.} Kinetic FID curves of different baselines over time.}
    \label{fig:long}
\end{figure}

\begin{table*}[t]
    \centering
    \caption{\textbf{Ablation Study.} \emph{paired} and \emph{unpaired} mean training with paired and unpaired data, respectively. \emph{MT} denotes fusion using multimodal Transformer \cite{li2021learn}. \emph{CLN} denotes fusion using conditional layer normalization \cite{kexuefm-7124}. \emph{Long} denotes the long-history motion encoder.}
    \label{tab:ablation_real}
    \begin{tabular}{lcccccccc} 
        \hline
         & \multicolumn{2}{c}{Quality} & \multicolumn{2}{c}{Diversity} & \multicolumn{2}{c}{Motion-Music Corr} &
         \multicolumn{1}{c}{User Study} \\
         &  $\text{FID}_{k} \downarrow$ & $\text{FID}_{g} \downarrow$ & $\text{Dist}_{k} \uparrow$ & $\text{Dist}_{g} \uparrow$ &  $\text{BeatAlign}\uparrow$ & $\text{Acc}_S\uparrow$ & Human Score$\uparrow$ \\
        \hline
        paired only MT & 81.32 & 26.38 & 4.71 & 4.00 & 0.219 & 0.14 & $2.25 \pm 0.29$\\
        paired only CLN & 61.47 & 21.54 & 5.58 & 4.41 & 0.220 & 0.51 & $2.40 \pm 0.31$\\
        paired w/o. Long & 33.23 & 16.25 & 5.19 & 5.61 & 0.222 & 0.52 & 2.96 $\pm$ 0.33 \\
        paired & \textbf{20.01} & \textbf{13.13} & \textbf{7.12} & 6.87 & 0.223 & \textbf{0.54} & \textbf{3.06} $\pm$ \textbf{0.35} \\
        unpaired & 25.10 & 15.63 & 6.81 & \textbf{7.04} & \textbf{0.223} & 0.50 & 3.03 $\pm$ 0.29 \\
        \hline
    \end{tabular}
\end{table*}
We conducted experiments to synthesize long-term dance. Because the test music clips in AIST++ are all within $30$ seconds, we repeat each sample to obtain music clip longer than $60$ seconds, and use them as test inputs. For each long clip, we determine an anchor per $3$ seconds and take an $1$-second generated motion clip centered on it. Then we calculate the $\eqword{FID}_k$ using the motion clips of each anchor. Finally, we get the $\eqword{FID}_k$ curve to measure the long-term generation performance of different baselines. As shown in Figure \ref{fig:long}, motions synthesized by our method have the lowest $\eqword{FID}_k$ score, and $\eqword{FID}_k$ remains stable over time. The model are difficulty to predict the correct subsequent dance of seed motion, so the $\eqword{FID}_k$ in the first few seconds is relatively high. Moreover, style consistency study of long-term generation was included in the supplementary materials.

\subsection{User Study} 
We organized a user study to evaluate the performance of our system qualitatively. The study was conducted with $11$ participants and consists of win rate scale and human score. For the win rate scale, each participant was requested to compare $40$ disordered pairs of motion and choose the better one in each pair. Each pair contains two motion sequences synthesized by our model and a baseline method, respectively. The final result is the probability that our model wins. The human score is a five-point Likert scale. Each participant was asked to grade the generated motions of each model from $1$ to $5$, with $1$ being the worst, and $5$ being the best.
Results in Table \ref{tab:analysis_real} show that our method surpasses the compared methods with at least $78.5\%$ wining rate. 

\subsection{Ablation Study}
We conducted ablation study to demonstrate the effectiveness of our proposed strategies. Results are shown in Table \ref{tab:ablation_real}, where \emph{MT} denotes fusion via multimodal Transformer \cite{li2021learn}, \emph{CLN} denotes fusion using conditional layer normalization, \emph{Long} means using long-history encoder, and \emph{paired} denotes applying paired learning scheme.

\subsubsection{Unpaired Learning Scheme}
We compared the results of different learning schemes (\emph{paired} and \emph{unpaired}) in Table \ref{tab:ablation_real}. Our proposed unpaired learning scheme can achieve comparable results even if only the unpaired data was used.

\subsubsection{Long-history Encoder}
Comparison of \emph{paired} and \emph{paired w/o. Long} in Table \ref{tab:ablation_real} showed that the long-history encoder can improve the quality and style classification accuracy of generated motions. Such results indicate that our long-history attention mechanism can alleviate the error accumulation problem caused by the autoregressive framework.

\subsubsection{Conditional Layer Normalization}
We chose two representative approaches of the multimodal Transformer (MT) and conditional layer normalization (CLN) to study the effects of different music context fusion strategies. We constructed a test setting that the model is only composed of a music encoder and a motion generator, and applied the paired training scheme. Then, the only variable is the fusion method. The comparison results of \emph{paired only MT} and \emph{paired only CLN} in Table \ref{tab:ablation_real} showed that CLN is superior than MT. Additionally, compared to MT, CLN can reduce the computational consumption. Assuming that the lengths of the input music and historical motion are $m$ and $n$ respectively, the computational complexity of the self-attention layer in MT-based Transformer is $O((m+n)^2)$, while that of the CLN-based Transformer is $O(n^2)$.

%% file: sections/6_conclusion.tex
\section{Conclusion}
\label{sec:conclusion}
In this paper, we present a 3D dance synthesis system that can generate realistic and style-consistent dance movements to accompany a piece of long music. For the lack of data, we propose an efficient unpaired learning scheme to take advantage of the easy-to-get unpaired data. For the synthesis of long-term dance, we devise a long-history attention mechanism based on choreography theory, which can perceive the choreographic structure from long historical motion and ensure style-consistent during long-term generation. Our system is comparable to strong baseline systems both objectively and subjectively, despite using only the unpaired data. The long-term dance generation experiment demonstrates the robustness of our system. Finally, we conduct detailed ablation studies that justify the effectiveness of our designs.

There are still many open topics for future exploration. First, identifying onsets as the music beats is not perfect. Onsets generally correspond to musical stresses, but in practice, dance beats may not always coincide with stress notes. Second, there are more to choreographic rules than the motion repeat constraint. For instance, identical bars in a musical phrase often correspond to symmetrical motions, which is called the \emph{mirror constraint} \cite{chen2021choreomaster}. Last, our unpaired learning scheme are scalable for larger unpaired datasets. Building a general pre-trained model based on a large-scale unpaired music-dance dataset is an interesting topic worth exploring.

\section{Acknowledgement}
\label{sec:acknowledgement}
This paper is partially supported by grants from the National Key Research and Development Program of China with Grant No. 2018AAA0101902 and the National Natural Science Foundation of China (NSFC Grant Number 62276002).

%% file: sections/7_appendix.tex
\clearpage
\appendix

\begin{figure*}[t]
    \centering
    \includegraphics[width=160mm]{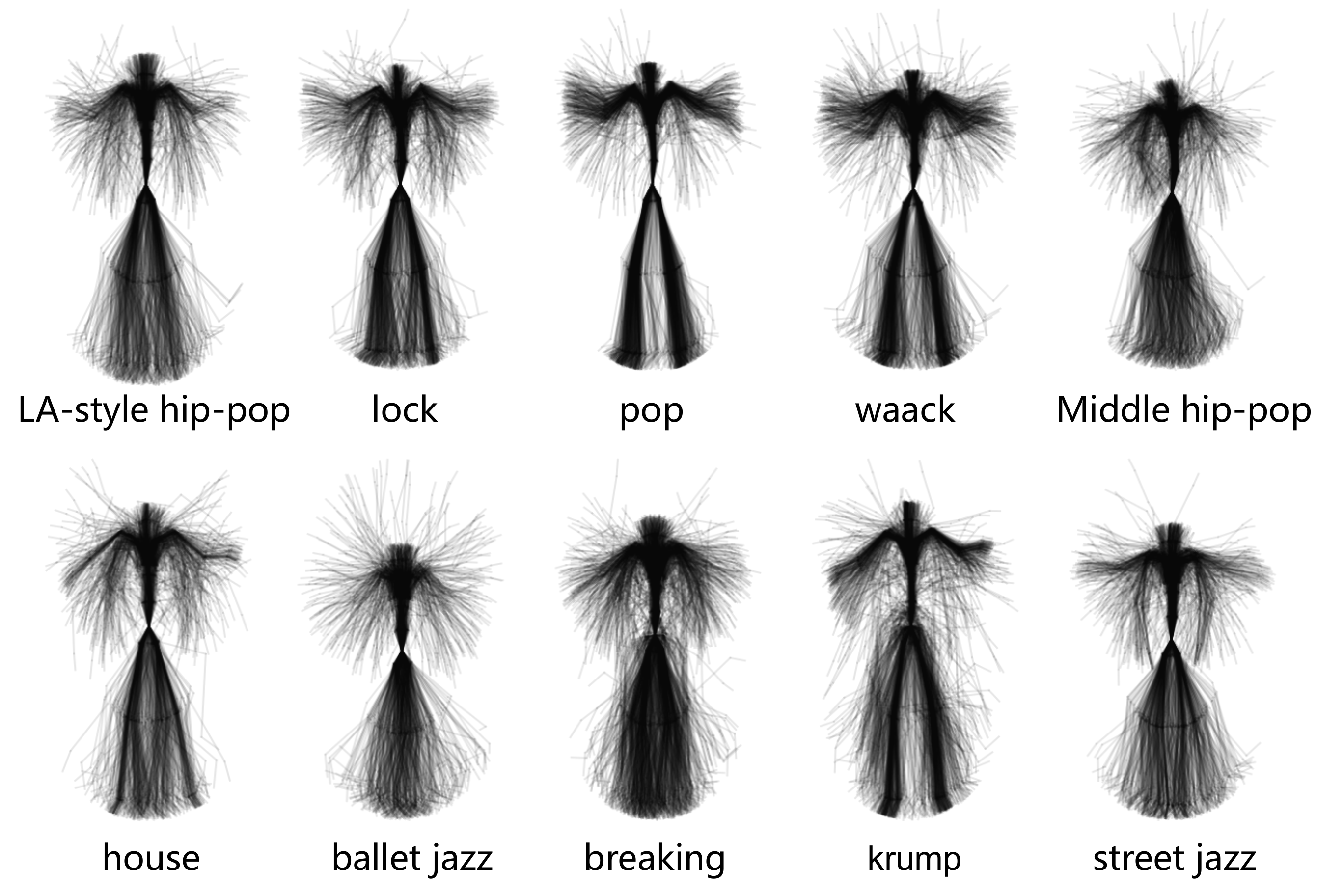} 
    \caption{\textbf{Pose Distributions of Generated Dance Motions on Different Styles of Music.} Different styles of dance are clearly distinguished, and the movements of each style are diverse.}
    \label{fig:styles}
\end{figure*}

\begin{figure*}[t]
    \centering
    \includegraphics[width=160mm]{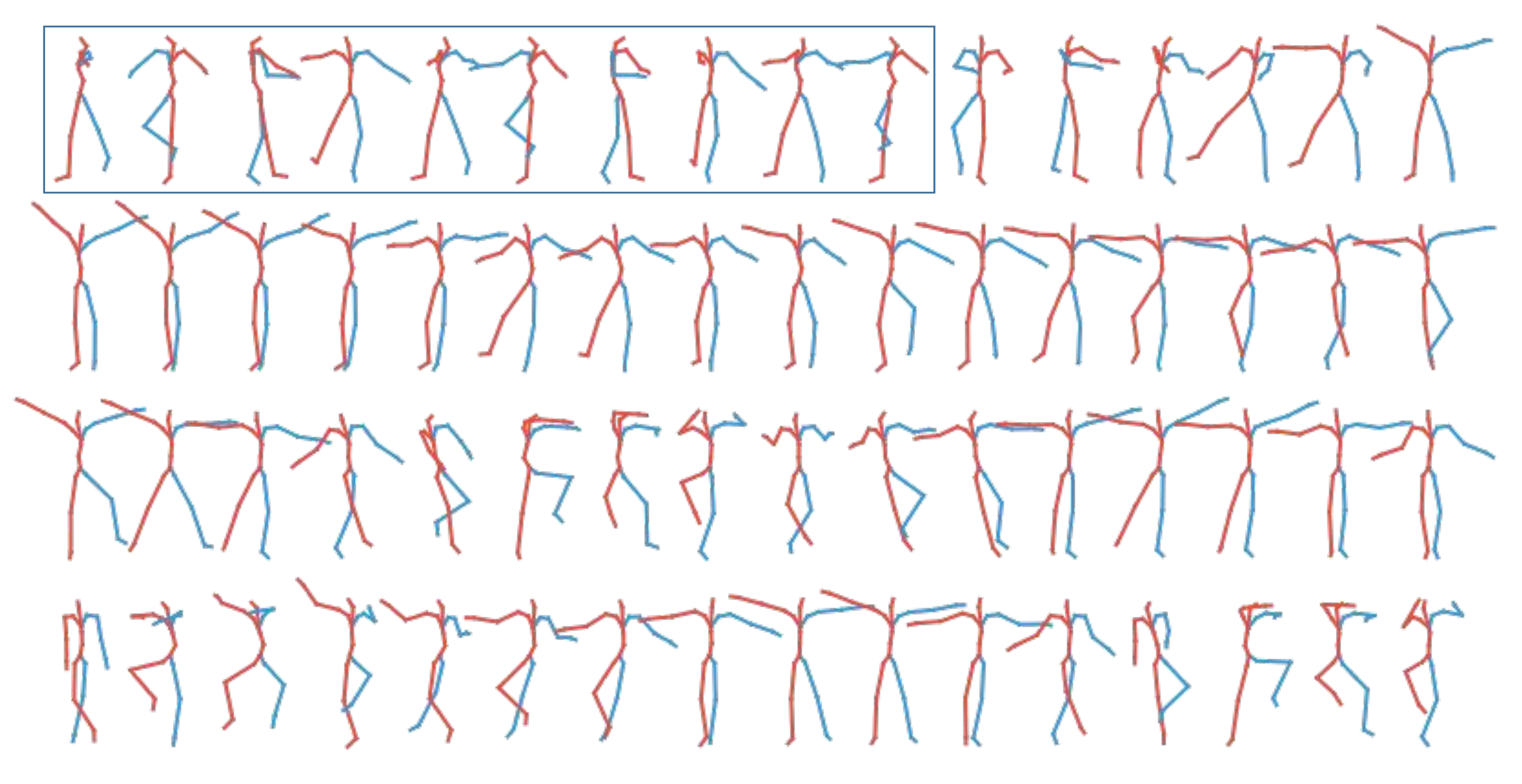} 
    \caption{\textbf{A Ballet Jazz Style Dance Synthesized by Our Method.} Poses in the blue box is the seed motion. The generated dance is realistic and style-consistent.}
    \label{fig:style_corr}
\end{figure*}

\begin{figure*}[t]
    \centering
    \includegraphics[width=160mm]{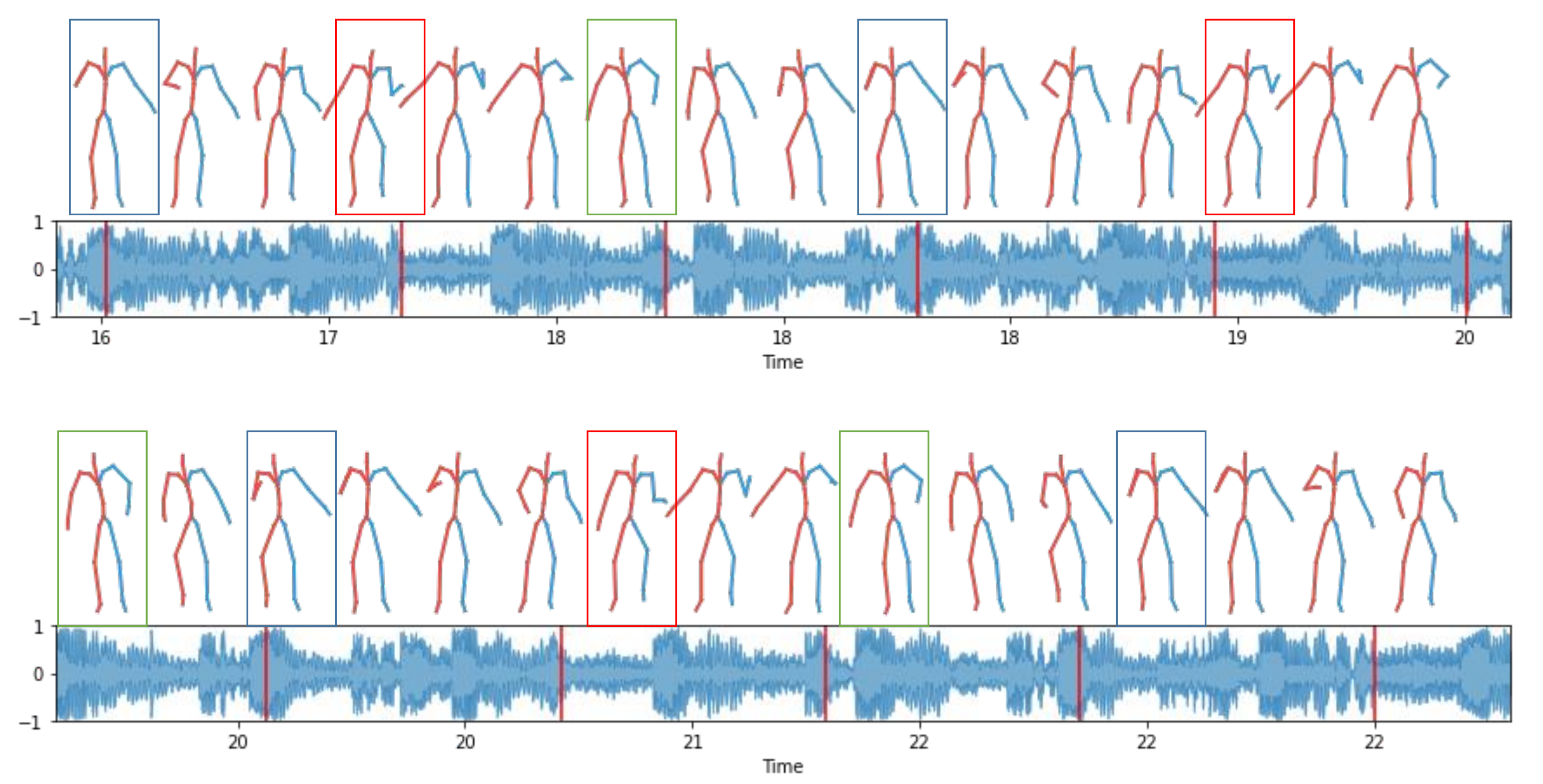} 
    \caption{\textbf{Visualization of Beat Matching.} A lock dance generated by our method. Red lines denote the musical beats, while kinetic beats are marked by boxes with different colors. The generated dance and music are well-matched in rhythm.}
    \label{fig:beat_corr}
\end{figure*}

\begin{figure*}[t]
    \centering
    \includegraphics[width=160mm]{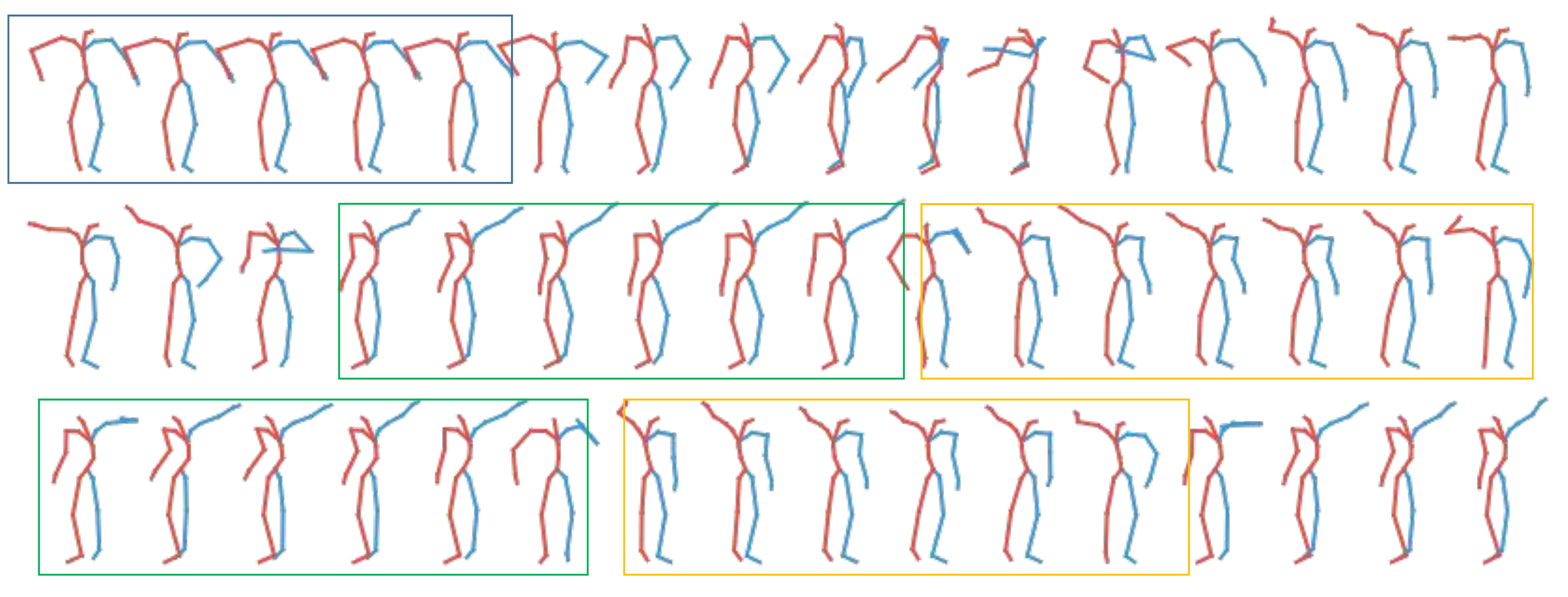} 
    \caption{\textbf{Perception of Choreographic Structure.} A house dance synthesized by our method, where the same color box frames the similar motion segment. The input music contains some cyclic phrases (e.g., chorus), and repeated musical phrases correspond to repeated generated movements, indicating that our long-history attention mechanism successfully perceives the structure of the music.}
    \label{fig:struct}
\end{figure*}